%% file: original.tex
\documentclass{article}

\usepackage{arxiv}
\usepackage{lipsum}

\usepackage[utf8]{inputenc}
\usepackage[T1]{fontenc}
\usepackage{hyperref}
\usepackage{url}
\usepackage{booktabs}       
\usepackage{amsfonts}
\usepackage{amsmath}
\usepackage{amssymb}
\usepackage{bbm}
\usepackage{multirow}
\usepackage{nicefrac}       
\usepackage{microtype}
\usepackage{graphicx}
\usepackage{caption, subcaption}
\usepackage[numbers]{natbib}
\usepackage{doi}

\title{Efficiently Visualizing Large Graphs}
\date{}

\author{
    Xinyu Li \\
	School of Art and Science \\
	New York University Shanghai \\
	Shanghai, China \\
	\texttt{\href{mailto:xl3665@nyu.edu}{xl3665@nyu.edu}} \\
	\And
    Yao Xiao \\
	School of Art and Science \\
	New York University Shanghai \\
	Shanghai, China \\
	\texttt{\href{mailto:yx2436@nyu.edu}{yx2436@nyu.edu}} \\
	\And
    Yuchen Zhou \\
	School of Art and Science \\
	New York University Shanghai \\
	Shanghai, China \\
	\texttt{\href{mailto:yz7258@nyu.edu}{yz7258@nyu.edu}} \\
}

\renewcommand{\headeright}{X.Li, Y.Xiao, Y.Zhou}
\renewcommand{\shorttitle}{Efficiently Visualizing Large Graphs}

\hypersetup{
pdftitle={A template for the arxiv style},
pdfsubject={q-bio.NC, q-bio.QM},
pdfauthor={David S.~Hippocampus, Elias D.~Striatum},
pdfkeywords={First keyword, Second keyword, More},
}

\begin{document}

\maketitle

\begin{abstract}
	\input{Sections/0_Abstract}
\end{abstract}


\section{Introduction}\label{introduction}
\input{Sections/1_Introduction}

\section{Method}\label{method}
\input{Sections/2_Method}

\section{Experiments}\label{experiments}
\input{Sections/3_Experiment_and_Result}

\section{Discussion}\label{discussion}
\input{Sections/4_Discussion}



\end{document}

%% file: Sections/0_Abstract.tex
Most existing graph visualization methods based on dimension reduction are limited to relatively small graphs due to performance issues.
In this work, we propose a novel dimension reduction method for graph visualization, called t-Distributed Stochastic Graph Neighbor Embedding (t-SGNE).
t-SGNE is specifically designed to visualize  cluster structures in the graph.
As a variant of the standard t-SNE method, t-SGNE avoids the time-consuming computations of pairwise similarity.
Instead, it uses the neighbor structures of the graph to reduce the time complexity from quadratic to linear, thus supporting larger graphs.
In addition, to suit t-SGNE, we combined Laplacian Eigenmaps with the shortest path algorithm in graphs to form the graph embedding algorithm ShortestPath Laplacian Eigenmaps Embedding (SPLEE).
Performing SPLEE to obtain a high-dimensional embedding of the large-scale graph and then using t-SGNE to reduce its dimension for visualization, we are able to visualize graphs with up to 300K nodes and 1M edges within 5 minutes and achieve approximately $10\%$ improvement in visualization quality.
Codes and data are available at \texttt{https://github.com/Charlie-XIAO/embedding-visualization-test}.

%% file: Sections/1_Introduction.tex
Graph data are widely used nowadays, and visualization of graph data is an important problem in various fields.
For example, interactions of users on social media websites can be represented by graphs with users as nodes and their following relations as edges.
Analysis on such social network graphs can give important information such as interpersonal ties, structural holes, and local online communities~\cite{chakraborty_application_2018}.
Various different types of methods for graph visualization have been proposed over the past few decades, many of which are reviewed and compared by Herman et al. in a survey on graph visualization and navigation~\cite{herman_graph_2000}.
Important techniques include force directed methods~\cite{kobourov_spring_2012} and spectral drawing methods~\cite{koren_drawing_2005}, etc.
Most of these techniques can be easily distracted by noises when plotting cluster structures of graphs.
They also run too slow or take too much space regarding large graph datasets consisting of over 10K nodes.
These severely limits their applicability on large graph data and real-world data where thousands of noises may exist.

To handle large-scale data, we found that a faster approach of recent years can be applied.
We can first compute a high dimensional embedding where each node in the original graph is marked by a vector.
After that we use dimension reduction methods to convert this embedding into a 2D layout.
There are various existing graph embedding methods, which can be found in the survey by Goyal and Ferrara~\cite{goyal_graph_2018}.
Existing popular dimension reduction methods include PCA~\cite{wold_principal_1987}, t-SNE~\cite{maaten_visualizing_2008} and UMAP~\cite{mcinnes_umap_2020}.
However, these methods are designed for data visualization and can not be directly applied to graph visualization. 
Moreover, although they can be combined with graph embedding to visualize graph, their runtime can still be improved. Also, their visualization quality is still a problem, especially regarding cluster structures.

In this paper, we focus on the simplest case of graph visualization: undirected, unweighted graphs without node attributes, and leave the generalization for future work.
Motivated by the objective of preserving the cluster structures of graphs, we propose ShortestPath and ShortestPath Laplacian Eigenmaps Embedding (SPLEE) that take into account graph theoretical distances and layout clusters more clearly.
Also, ShortestPath is a fast algorithm that can deal with larger datasets.
We also put forward a fast and cluster-preserving dimension reduction technique called t-SGNE based on the original t-SNE, which takes advantage of the graph neighbor structures.
Finally, in order to compare the results of different methods on different datasets, we propose two quantitative measures for testing, including Normalized Mutual Information (NMI) and Aesthetic Quality (AQ).
These measures respectively evaluate the clustering accuracy and how well clusters are distributed in layout.

\paragraph{Organization.}
In the remaining part of this section, we outline the previous works related to our work. In Section~\ref{method}, we discuss details of our methods, including graph embedding methods, dimension reduction methods and quantitative testing standards. In Section~\ref{experiments}, we present the testing results of both previous methods and our methods on a variety of datasets, regarding visualization quality and running time. We also provide the repository of our codes on GitHub.\footnote{https://github.com/Charlie-XIAO/embedding-visualization-test}
In Section~\ref{discussion}, we discuss possible future improvements on our work. 

\subsection{Related Work}
\input{Sections/1_1_Related_Work}

\subsection{Our Contribution}
\input{Sections/1_2_Our_Contribution}

%% file: Sections/1_1_Related_Work.tex
\subsubsection{Graph Layout}
\input{Sections/Related_Work_Subsections/1_1_1_Graph_Layout}

\subsubsection{t-SNE}
\input{Sections/Related_Work_Subsections/1_1_2_t-SNE}

\subsubsection{Graph Embedding}
\input{Sections/Related_Work_Subsections/1_1_3_Graph_Embedding}

%% file: Sections/Related_Work_Subsections/1_1_1_Graph_Layout.tex
A layout of a graph is a two dimensional embedding of the graph, where each node is assigned a coordinate on the 2D plane for visualization purpose. 
There are mainly three types of methods to compute the layout of a graph: force-directed methods, spectral methods, and dimension reduction methods.

\begin{itemize}
    \item 
    \textbf{Force directed methods} model nodes as particles that repel each other, and edges as springs that connect the particles. 
    Thus, it can compute the graph layout by simulating the corresponding particle-spring system and minimizing the energy of the system~\cite{kobourov_spring_2012}.
    It can produce aesthetically pleasing results with less edge crossing and uniform distribution of nodes. 
    However, the optimization of this complex system is time-consuming and computation-intensive, which means we cannot directly apply force directed method to graphs of large scale. 
    Moreover, the aesthetic criteria used in force directed method fail to reflect how the cluster structure of the graph is represented in the layout. In fact, empirically, for a graph with more than 10K nodes, force-directed methods usually produce a ``hairy ball'' with no identifiable clusters.

    \item
    \textbf{Spectral methods} use eigenvectors of some matrices related to the graph as the graph layout.
    Compared with force directed methods, spectral methods are much faster, most of which run quadratic time.
    Also, spectral methods are deterministic, presenting an exact mathematical formula that draws the layout.
    The most famous spectral drawing method is to compute the lowest eigenvectors of the Laplacian matrix of the graph~\cite{koren_drawing_2005}.
    The correctness of this method can be verified with an optimization problem.
    Spectral drawing methods tend to place each node at the centroid of its neighbors with some deviation.
    This preserves the graph-theoretical distance between nodes pretty well.
    However, this may also result in lots of crossings between edges and ambiguity of the borders of clusters.
    Different clusters may mix up a lot for large graphs.

    \item
    \textbf{Dimension reduction methods} have a wide range of usage. 
    They can be applied to graph drawing with certain modifications.
    Dimension reduction methods aim to project the given high dimensional data to lower dimension while preserving some form of information of the original high dimensional data (typically represented by an objective function).
    To apply dimension reduction method to graph drawing, one can first compute a high dimensional embedding of the graph, then reduce the dimension of the embedding to two while minimizing some form of difference between the high and low dimensional embedding. 
    The output can be treated as a graph layout.
    For each node, a typical choice of high dimensional embedding would be its graph-theoretic distance to all the nodes in the graph.
    Pivot MDS, proposed by Brandes and Pich, is a classic example~\cite{kaufmann_eigensolver_2007}. 
    It first samples some pivot nodes and uses the distances of each node to these pivots as high dimensional embeddings.
    Then, it applies Multi-Dimensional Scaling (MDS) to reduce the dimension to two, where the objective function is a so-called stress function that measures the discrepancy between the pairwise graph-theoretic distance in the graph and the pairwise Euclidean distance on 2D plane.
    Methods based on dimension reduction are suitable for the visualization of large-scale graph because there are various approximations of the objective functions of dimension reduction methods. 
    However, layouts produced by dimension reduction methods tend to be less aesthetic compared to those produced by force-directed methods.
\end{itemize}

%% file: Sections/Related_Work_Subsections/1_1_2_t-SNE.tex
t-SNE is a nonlinear, statistical model for dimension reduction. 
It aims to map some known high dimensional data to low dimension in a way that preserves the neighbor structure in a probabilistic way, whereas traditional dimension reduction methods like MDS preserves the more intuitive distance structure. 
Loosely speaking, t-SNE assumes that similarity between points are captured by their distance in high dimensional space. 
Thus, a point is similar to its neighbors and dissimilar to points far away.
The low dimensional embedding is constructed so that, in the low dimensional space, points that are similar in high dimensional space are closer and dissimilar points are farther apart with high probability.

The primary use of t-SNE is to visualize data in high dimensional space, thus a common choice of the low dimensional space is $\mathbb{R}^2$. 
Compared to previous data visualization methods, t-SNE can preserve both local structure and global structure such as clusters at different scales~\cite{maaten_visualizing_2008}.

To apply to large dataset, t-SNE uses an approximation based on random walk on neighborhood graph, where the neighborhood graph is constructed from the high dimensional embedding. Although this enable the application of t-SNE to larger dataset, the computation of neighborhood graph creates a computational bottleneck and restrict t-SNE to dataset of size about 100K~\cite{amid_trimap_2022}.

%% file: Sections/Related_Work_Subsections/1_1_3_Graph_Embedding.tex
Graph embedding is a powerful method for reducing the dimensions of graph data while preserving certain graph structures.
More specifically, a graph embedding is a mapping that maps each node to a representation vector, whose dimension is much smaller than the number of nodes.
In an effective graph embedding method, the representation vectors of nodes within the same community should be similar, so that closely-related nodes tend to lie close to each other after dimension reduction as well.
There are various types of graph embedding methods.

DeepWalk~\cite{perozzi_deepwalk_2014} is a random walk based embedding method.
It applies truncated random walks to walk through the nodes and obtain sampling.
A truncated random walk starts at a certain node and randomly visits one of its neighbors and so on, until the length of the walk reaches a default value.
This describes the cooccurrence relations of nodes, which is the key of this method.
DeepWalk then uses word2vec~\cite{church_word2vec_2017} to create the representation vectors of each node.
Word2vec is a common method for word embedding in NLP, which learns the cooccurrence relations among words from sentences and vectorize each word.
DeepWalk is hence able to vectorize each node by learning their cooccurrence relations.

Node2Vec~\cite{grover_node2vec_2016} is another random walk based embedding method.
It is similar to DeepWalk but samples random walks differently.
It introduces two parameters, which controls the probability of performing a breadth-first search or a depth-first search when randomly choosing the next node to visit.
Breadth-first search better records the similarity of closely-related nodes.
Depth-first search may preserve some global structures of the original graph.

Laplacian Eigenmaps~\cite{belkin_laplacian_2003} is a different type of graph embedding method.
It constructs relations between nodes from a local perspective.
Since the objective is to keep the closely-related nodes close to each other in the lower-dimensional space, Laplacian Eigenmaps tries to minimize $\sum_{i,j}\lVert y_i-y_j\rVert$, where $y_i$ and $y_j$ are data points in the lower-dimensional space.
The problem becomes minimizing the trace of $Y^\mathrm{T}LY$ after some transformation, where $L=D-A$ is called the unnormalized Laplacian matrix, $A$ is the adjacency matrix, and $D$ is a diagonal matrix with entries $D_{ii}=\sum_jA_{ij}$.
By trace derivative law, $LY=-DY\Lambda$ gives the optimized result.
This can be rewritten as the generalized eigenvalue problem $Ly=\lambda Dy$.
Hence, Laplacian Eigenmaps first computes the eigenvalues and eigenvectors of the Laplacian matrix of the graph, then takes the $d$ eigenvectors corresponding to the $d$ smallest nonzero eigenvalues as the $m$-dimensional output ($d\ll |V|$ is the dimension of the high dimensional embedding).

Geometric Laplacian Eigenmaps~\cite{torres_glee_2020} is similar to Laplacian Eigenmaps while taking the eigenvectors corresponding to the $m$ largest nonzero eigenvalues.
The intuition is that these correspond to the best approximation to the Laplacian through singular value decomposition.

%% file: Sections/1_2_Our_Contribution.tex
Our work involves three parts: graph embedding using ShortestPath (SP) and ShortestPath Laplacian Eigenmaps Embedding (SPLEE), dimension reduction using t-SGNE, and two quantitative testing standards, Normalized Mutual Information (NMI) and Aesthetic Quality (AQ). Using SPLEE for graph embedding first and then using t-SGNE to perform dimension reduction, we show that we can visualize graphs with up to 300K nodes and 1M edges within $5$ minutes and achieve approximately $10\%$ improvement in visualization quality.

%% file: Sections/2_Method.tex
The problem of graph drawing can be formulated as follows.
Let $G=(V, E)$ be an undirected, unweighted graph where $|V|=n$ (for more general types of graphs, see Section~\ref{discussion}).
Graph Drawing defines a function $GD: G \mapsto Y$, where $Y=\{y_i \in \mathbb{R}^2 \: | \: i=1,\cdots,n\} \subset \mathbb{R}^2$ is the two dimensional embedding of the graph $G$.

There are two steps to apply the method of dimension reduction to graph drawing.
The first step is Graph Embedding ($GE$).
Given a graph $G$, we want to find a function $GE: G \mapsto X$ where $X = \{x_i \in \mathbb{R}^d \: | \: i=1,\cdots,n\} \subset \mathbb{R}^d$ is the high dimensional embedding of the graph $G$ and $d \gg 2$.
Next, we will perform Dimension Reduction ($DR$) on $X$, that is, to apply a function $DR: X \mapsto Y$ on $X$. Thus, we have the composition
\begin{align}
GD = DR \circ GE.
\end{align}

\subsection{t-SNE}
\input{Sections/Method_Subsections/t-SNE}

\subsection{t-SGNE}
\input{Sections/Method_Subsections/t-SGNE}

\subsection{ShortestPath}
\input{Sections/Method_Subsections/ShortestPath}

\subsection{SPLEE}
\input{Sections/Method_Subsections/SPLEE}

\subsection{Normalized Mutual Information}
\input{Sections/Method_Subsections/Normalized_Mutual_Information}

\subsection{Aesthetic Quality}
\input{Sections/Method_Subsections/Aesthetic_Quality}

%% file: Sections/Method_Subsections/t-SNE.tex
t-SNE is a $DR$ method. We will first state the original t-SNE algorithm, then describe its approximation based on random walk on neighborhood graph, as is done in the original paper.
The original t-SNE consists of three parts: constructing a probability distribution $\mathcal{P}$ in $\mathbb{R}^d$ and approximate $\mathcal{P}$ with probability distribution $\mathcal{Q}$ in $\mathbb{R}^2$.

First, for $x_i, x_j \in X$, we can compute the probability $p_{j|i}$ that $x_i$ would pick $x_j$ as its neighbors as follows:
\begin{align}
p_{j|i} = \frac{\exp(-\lVert x_i-x_j\rVert^2/2\sigma_i^2)}{\sum_{k \not = i} exp(-\lVert x_i-x_k\rVert^2/2\sigma_i^2)} \cdot \mathbbm{1}_{\{i \not = j\}},
\end{align}
where $\chi_A$ is the indicator function of set $A$.
Then, we can define the distribution $\mathcal{P}$ as
\begin{align}
p_{ij} = \frac{p_{j|i}+p_{i|j}}{2n},
\end{align}
where $\sigma_i$ is chosen using a binary search so that the perplexity $\mathrm{Perp}(\mathcal{P}_i) = 2^{H(\mathcal{P}_i)}$ equals to a user-defined value ($\mathrm{Perp}(\mathcal{P}_i)$ is typically between $5$ and $50$, scikit-learn choose$30$ as default), and $H(\mathcal{P}_i) = -\sum_j p_{j|i}\log p_{j|i}$ is the Shannon entropy as is suggested in SNE~\cite{hinton_stochastic_2002}.

Next, we construct $\mathcal{Q}$ in a similar way, except we use the Student t-distribution instead of Gaussian distribution:
\begin{align}
q_{ij} = \frac{(1+\lVert y_i-y_j\rVert^2)^{-1}}{\sum_k\sum_{l \not = k}(1+\lVert y_k-y_l\rVert^2)^{-1}} \cdot \mathbbm{1}_{\{i \not = j\}}.
\end{align}

The Student t-distribution uses a degree of freedom equal to 1, which results in a Cauchy distribution. This distribution is heavy-tailed (infinite first moment), which helps alleviate the crowding problem of Stochastic Neighbors Embedding (SNE), that data points are all mapped to the center becoming indistinguishable~\cite{maaten_visualizing_2008}.

To find $\mathcal{Q}$ that best approximates $\mathcal{P}$, we use the KL divergence of $\mathcal{P}$ from $\mathcal{Q}$ as an objective:
\begin{align}
\min_{Y} \mathrm{KL}(\mathcal{P}\Vert \mathcal{Q}) = \sum_{i,j} p_{ij} \log\left(\frac{p_{ij}}{q_{ij}}\right).
\end{align}

To apply t-SNE to larger datasets, we use an approximation of $P$ based on neighborhood graph and random walk.
First, we construct a neighborhood graph $G_{knn} = G_{knn, X}$ from $X$, where $x_i$ is linked to its first $k$ nearest neighbors, denoted as $N(i)$.
We perform a fixed large number of random walks of fixed length on the graph for each node, where the probability of transiting from $x_i$ to $x_j$ is proportional to $\exp{\lVert x_i-x_j\rVert^2}$.
From the random walks, we can construct an approximation of $p_{j|i}$ as follows:
\begin{align}
p_{j|i} = \frac{\textrm{number of random walks from $i$ to $j$}}{\textrm{number of random walks starting from $i$}}.
\end{align}
The rest is the same as what we stated above.

\paragraph{Time complexity.}
t-SNE mainly involves three steps.
First, the construction of the $G_{knn,X}$ involves the computation of pairwise Euclidean distances, which has a time complexity of $O(d|V|^2)$.
The reason is that for each node, we need to further compute and rank its distance to the rest of the nodes, and select the first $k$ nearest nodes to construct the neighborhood graph.
Second, the simulation of random walks runs $O(|V|)$ time.
Finally, the optimization of $\mathrm{KL}(\mathcal{P}\Vert \mathcal{Q})$ has a time complexity of $O(|V|^2)$ as is suggested in the original article~\cite{maaten_visualizing_2008}.

%% file: Sections/Method_Subsections/t-SGNE.tex
t-SGNE is a simple yet effective modification of t-SNE that can be applied to graph data with high dimensional embeddings.
Given a graph $G$ and its high dimensional embedding $X$, we perform t-SNE with nearest neighbor approximation, except that the neighborhood graph is constructed from $G$ instead of $X$, i.e.
\begin{align}
G_{knn} = G_{knn, G},
\end{align}
where $v_i, v_j \in G$ are connnected in $G_{knn, G}$ if $v_i$ is of the first $k$ nearest neighbors of $v_j$ on graph $G$, where the nearest neighbor is computed by a breadth-first search of $k$ steps.

To justify this simple modification, we need to answer a nontrivial question: is $G_{knn, G}$ a valid substitute for $G_{knn, X}$?
The answer depends on our choice of $GE$ method.
The goal of t-SNE is to map points closer (farther) in $\mathbb{R}^d$ to closer (farther) positions in $\mathbb{R}^2$.
When applied to large datasets, $G_{knn, G}$ with random walk approximates the distance relation of points in $\mathbb{R}^d$. 
If $x_i$ and $x_j$ are closer in $\mathbb{R}^d$, a random walk starting from $x_i$ is more likely to reach $x_j$, which results in larger $p_{j|i}$.
However, points closer in $G$ are not necessarily closer in $\mathbb{R}^d$, and vice versa.
For example, Structural Deep Network Embedding (SDNE) is a $GE$ method that captures structural similarity instead of distance/neighbor relations between nodes in a graph~\cite{hinton_stochastic_2002}.
Nodes are assigned to closer positions in $\mathbb{R}^d$ not because they are closer in the graph but because they have similar structural properties (e.g. both are central to a cluster, both are hubs between two clusters). 
In this case, simply comparing the ratio of random walk paths from $v_i$ to $v_j$ will not reflect the structural similarity and dissimilarity between these nodes.

To judge if t-SGNE is suitable, we need to ask another question: when we map a graph $G$ to $X \subset \mathbb{R}^d$, what kind of points should be closer in the high dimensional space $\mathbb{R}^d$? 
Typically, t-SGNE can only be combined with $GE$ methods that maps nodes of small geodesic distance (shortest path length) to closer embeddings.
For $GE$ methods that aim to preserve more complex structures like structural similarity, $G_{knn, G}$ is not a good approximation of $G_{knn,X}$. 
In other words, with an appropriate $GE$ method, t-SGNE is suitable for visualizing the cluster structures of graph, which is captured by the geodesic distance relation in $G$.

\paragraph{Time complexity.}
The construction of $G_{knn, G}$ has a time complexity of $O(k|V|)$ since for each node, we need to perform a $k$-step BFS to determine its $k$ neighbors.
Compared with t-SNE which involves a quadratic-time computation of $G_{knn,X}$, t-SGNE only requires linear time to construct the neighborhood graph.
The rest is the same as t-SNE.

%% file: Sections/Method_Subsections/ShortestPath.tex
ShortestPath is a $GE$ method that gives graph embedding of each node based on its shortest path length to certain target nodes.
The motivation that we introduce such method is that we want the embedding result to have association with the distance between nodes in the original data, which is mostly represented by the shortest path lengths between nodes. 

First, we pick the target nodes.
Let $d$ be the dimension of the high dimensional embedding.
Through experiments, we found that randomly choosing $d$ targets shows the overall best result.

Then, to calculate the length of shortest paths between each node in the graph and each target in $X$, we apply BFS starting from each node in $X$.
To minimize the negative effect of nodes that are far apart in the final plot, we also apply a threshold $l_0$ here.
If the length of the shortest path is less than $l_0$, we use it in the embedding.
Otherwise, we regard the length as $l_0+1$.
The default value of $l_0$ is $\sqrt{|E|}$ where $|E|$ is the number of edges in the graph.
Such a threshold also reduces the number of computations thus increasing efficiency.

Hence, the ShortestPath embedding is defined as follows:
we first obtain an embedding matrix
\begin{align}
    E_{ij}=
    \begin{cases}
        d_{ij}, & \text{if $d\leq l_0$}, \\
        l_0+1, & \text{if $d>l_0$ or path does not exists between $v_i$ and $x_j$},
    \end{cases}
\end{align}
where $d_{ij}$ represents the shortest path length between the nodes $v_i\in V$ and $x_j\in X$. Each row vector is then taken as the embedding vector of node $v_i$.

\paragraph{Time complexity.}
Since BFS is applied $d$ times in total, and in each round it traverses each edge at most once, the total time complexity of ShortestPath is $O(d|E|)$.

%% file: Sections/Method_Subsections/SPLEE.tex
ShortestPath Laplacian Eigenmaps Embedding (SPLEE) is a $GE$ method that combines ShortestPath and Laplacian Eigenmaps.
The motivation is to take into account graph-theoretically distances rather than just considering node connections when doing spectral embedding.

The original version of Laplacian Eigenmaps~\cite{belkin_laplacian_2003} can be applied on more general types of high dimensional data.
However, they need to be transformed to weighted graphs before applying the method.
Links exist if corresponding data points are close to each other, and weights are determined by some functions on the Euclidean distances between data points which will be mentioned later soon.
However, when applying Laplacian Eigenmaps on graph data, no types of distances are taken into consideration.
Hence, SPLEE uses shortest path lengths between nodes to make up for this.

The algorithm involves two main steps.
The first step is to obtain a special distance matrix $W$.
As in the original Laplacian Eigenmaps, a heat kernel is applied to the Euclidean distances to approximate the Gaussian~\cite{belkin_laplacian_2003}.
Similarly, SPLEE applies the heat kernel to the shortest path lengths between nodes.
Furthermore, $W_{ij}=0$ if the shortest path length between the nodes $v_i$ and $v_j$ are beyond some threshold $l_0$ since nodes that are far apart should not be linked as in the original version of Laplacian Eigenmaps.
Hence, the distance matrix $W$ is defined as:
\begin{align}
    W_{ij}=
    \begin{cases}
        \exp(-\epsilon d_{ij}^2), & \text{if $d\leq l_0$}, \\
        0, & \text{if $d>l_0$ or path does not exists between $v_i$ and $v_j$},
    \end{cases}
\end{align}
where $d_{ij}$ represents the shortest path length between $v_i$ and $v_j$.
According to experimental results, $\epsilon$ is recommended to be around 5.0 to 7.0.
The threshold $l_0$ can be chosen as the same default value $\sqrt{|E|}$ as in ShortestPath.
The shortest path lengths are computed by BFS for unweighted graphs (or by Dijkstra's Algorithm for weighted graphs).

The second step is to compute eigenvectors to use as the high dimensional node embeddings. We generalize the original definition of Laplacian matrix $L$ as:
\begin{align}
L=D-W,
\end{align}
where $W$ is the distance matrix above, and $D$ is a diagonal matrix with entries $D_{ii}=\sum_jW_{ij}$.
Then the eigenvectors corresponding to the smallest $d$ eigenvalues of the generalized Laplacian matrix $L$ are taken as the $d$-dimensional embedding ($d\ll|V|$).

\paragraph{Time complexity.}
SPLEE uses the same technique of computing shortest path lengths as ShortestPath, which takes $O(d|E|)$ time. To compute the lowest $d$ eigenvectors of the Laplacian matrix, the state-of-art algorithm takes $O(k|V|^2)$ time, where $k$ is the number of iterations. Hence, SPLEE runs a total time complexity of $O(d|E|+k|V|^2)$.

%% file: Sections/Method_Subsections/Normalized_Mutual_Information.tex
This is a measure for clustering accuracy of the 2D layout.
Normalized Mutual Information (NMI) is a measure that compares two network partitions.
It ranges from 0 to 1, and the smaller the NMI, the more similar the two partitions.
Here we do clustering for both the original graph and the 2D graph layout, and compare the clustering results.
There are two main steps to achieve this.

First, we use Louvain's algorithm~\cite{blondel_fast_2008} to cluster the original graph.
In this step, we output the clustering label for each node as well as the number of clusters created.
One important reason to choose this algorithm is that it works directly on graph data structures and does not require the number of clusters as a parameter.
In this way, we can obtain the optimal clustering result for the original graph to use as a standard community partition.

Second, we apply kNN clustering algorithm on the 2D graph layout, with the $k$ chosen as the output number of clusters in the previous step.
Hence, we can keep in accordance the number of clusters between the graph and the layout.
In this way, the two sets of clustering labels have the same number of distinct values, thus the result will be more intuitive and accurate.
Now we compute the NMI as follows:
\begin{align}
\mathrm{NMI}(PR\,;TR)=\frac{2\cdot I(PR\,;TR)}{H(PR)+H(TR)},
\end{align}
where $PR$ and $TR$ denote the set of labels of the original graph and the 2D layout respectively, $H(\cdot)$ denotes the entropy, and $I(\cdot\,;\cdot)$ denotes the mutual information between the two arguments.

%% file: Sections/Method_Subsections/Aesthetic_Quality.tex
This is a measure for the aesthetic quality focusing on cluster structures of the 2D layout.
It examines whether clusters are clearly divided.
It ranges from 0 to 1, and the bigger the value, the better clusters are distributed in the layout.
The basic idea is to divide the plot into $k\times k$ grids.
For each grid, we calculate the portion that nodes of each cluster take up and we compare it with a user-defined threshold $p$.
If the result is larger than $p$, we regard the grid a good grid with a clear dominating cluster, and otherwise we ignore it.
Finally, we give an overall result by calculating the ratio of good grids.

Specifically, the x-length of each grid is given by
\begin{align}
    x_0=\frac{X_{\max}-X_{\min}}{k},
\end{align}
where $X_{\max}$ is the largest value along the x-axis of the point position in the layout and $X_{\min}$ is the smallest.
Similarly, we define the y-length of each grid as
\begin{align}
    y_0=\frac{Y_{\max}-Y_{\min}}{k},
\end{align}

Thus, we can easily know the bounds for each grid.
Now, by going over the position of every node in the layout, we can specify which grid it belongs to.
Then by calculating the portion each kind of nodes takes up in the grid, we can examine whether it is a good grid. The result of this measure $AQ$ is thus given by
\begin{align}
    AQ=\frac{\text{number of good grids}}{k^2}.
\end{align}

%% file: Sections/3_Experiment_and_Result.tex
In this section we present experimental analyses on our methods.
We first compare different combinations of $GE$ and $DR$ methods to compare their visualization quality, then we take a specific combination (ShortestPath and t-SGNE) for testing on larger graph datasets.
Testing codes, datasets, and part of the experimental results are available on GitHub.\footnotemark[1]

\subsection{Visualization quality of different combinations}\label{experiment1}
\input{Sections/Experiments/01}

\subsection{Runtime of t-SNE and t-SGNE}\label{experiment2}
\input{Sections/Experiments/02}

\subsection{Visualization of large graphs}\label{experiment3}
\input{Sections/Experiments/03}

%% file: Sections/Experiments/01.tex
We evaluate our new $GE$ methods ShortestPath and SPLEE and new $DR$ method t-SGNE.
We also compare with the existing $GE$ method DeepWalk and $DR$ method t-SNE as a baseline.
This experiment focuses on comparing the visualization quality of different combinations of methods, hence only small graph datasets (< 10K nodes) are used.
Larger datasets will be tested in the next experiment.
We experiment on three different datasets from Network Data Repository~\cite{ryan_network_nodate}.

\begin{itemize}
    \item \textsc{EmailUniv} is a network of a small collection of emails sent universally.
    It has 1133 nodes and 5451 edges, with an average degree of 9.
    It has clear cluster structures representing frequent sending and receiving of emails within certain communities.
    
    \item \textsc{Wiki} is a Wikipedia-based network constructed from different Wikipedia categories.
    It is a originally a labeled graph, but we will not take its labels into account.
    It has 2405 nodes and 12761 edges, with no clear borders of cluster structures due to the high interpenetration between the topics collected in this dataset.
    
    \item \textsc{Lastfm} is a heterogeneous network of user relations of the music website Last.fm.
    Each node represents an Asian user of Last.fm, and the edges represent the following relations between them.
    The network consists of 7624 nodes and 27806 edges, with a small density of 0.001.
    Cluster structures are clear in this network, representing the online communities of Asian Last.fm users.
\end{itemize}

\begin{table}[t]
    \centering
    \begin{tabular}{|c|c|c|c|c|c|c|}
        \hline
        \textbf{Dataset} & \multicolumn{2}{|c|}{\textbf{Method ($GE$ / $DR$)}} & \multicolumn{2}{|c|}{\textbf{Time ($GE$ / $DR$)}} & \textbf{NMI} & \textbf{AQ} \\
        \hline
        \multirow{6}{*}{\textsc{EmailUniv}} & DeepWalk & t-SNE & 10.102 & 5.715 & 0.4951 & 0.50 \\
        & ShortestPath & t-SNE & 0.394 & 5.848 & 0.2112 & 0.30 \\
        & SPLEE & t-SNE & 3.231 & 5.916 & \textbf{0.6106} & \textbf{0.57} \\
        & DeepWalk & t-SGNE & 9.530 & 6.219 & 0.4917 & \textbf{0.57} \\
        & ShortestPath & t-SGNE & 0.298 & 6.442 & 0.4917 & \textbf{0.57} \\
        & SPLEE & t-SGNE & 3.369 & 6.226 & 0.4917 & \textbf{0.57} \\
        \hline
        \multirow{6}{*}{\textsc{Wiki}} & DeepWalk & t-SNE & 24.279 & 17.410 & 0.6146 & \textbf{0.59} \\
        & ShortestPath & t-SNE & 0.883 & 15.690 & 0.3627 & 0.24 \\
        & SPLEE & t-SNE & 20.382 & 16.925 & \textbf{0.6147} & \textbf{0.59} \\
        & DeepWalk & t-SGNE & 21.433 & 14.974 & 0.5988 & 0.56 \\
        & ShortestPath & t-SGNE & 0.817 & 15.273 & 0.5883 & 0.53 \\
        & SPLEE & t-SGNE & 17.759 & 14.508 & 0.5944 & 0.50 \\
        \hline
        \multirow{6}{*}{\textsc{Lastfm}} & DeepWalk & t-SNE & 65.456 & 41.130 & 0.6822 & \textbf{0.67} \\
        & ShortestPath & t-SNE & 2.881 & 37.680 & 0.5210 & 0.56 \\
        & SPLEE & t-SNE & 343.962 & 40.694 & \textbf{0.6894} & \textbf{0.67} \\
        & DeepWalk & t-SGNE & 71.484 & 44.128 & 0.6746 & 0.62 \\
        & ShortestPath & t-SGNE & 2.920 & 44.354 & 0.6746 & 0.62 \\
        & SPLEE & t-SGNE & 343.272 & 44.723 & 0.6746 & 0.62 \\
        \hline
    \end{tabular} \\ \vskip .3cm
    \caption{\label{exp_01} Comparisons on running time (in seconds), NMI, and AQ scores of different combinations of $GE$ and $DR$ methods. The running time is experimented on a single machine with 2.3 GHz Intel Core i7 CPU and 16 GB of memory. While the running time gives an indication on the applicability of the methods on large graphs, the main focus of this experiment is on NMI and AQ scores which represent the visualization quality.}
\end{table}

\paragraph{Running time.}
As for running time of the $GE$ methods, ShortestPath takes the lead, with a linear runtime with respect to the number of edges.
DeepWalk takes longer than ShortestPath since it needs to train the random walk model, but it is less affected by the increase of graph size.
Due to the quadratic runtime of SPLEE, it runs much slower than the other two $GE$ methods when the size of graph gets bigger, so it may not be a good choice for large graphs consisting of 10K nodes and more.
As for the running time of the $DR$ methods, t-SNE and t-SGNE do not differ much, since the size of the high dimensional embedding is set to be 128 here.
However, if the size of the high dimensional embedding get larger, t-SGNE may outperform t-SNE since it takes advantage of graph neighbors rather than computing pairwise similarity of nodes.

\begin{figure}[htb]
    \centering
    \begin{subfigure}{0.4\textwidth}
        \includegraphics[width=\linewidth]{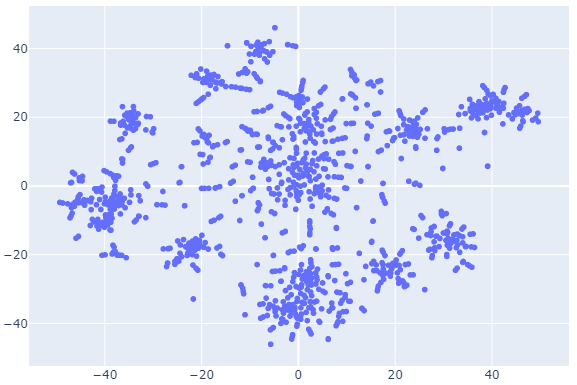}
        \caption{SPLEE + t-SNE}
        \label{fig:aa}
    \end{subfigure}\hfil
    \begin{subfigure}{0.4\textwidth}
        \includegraphics[width=\linewidth]{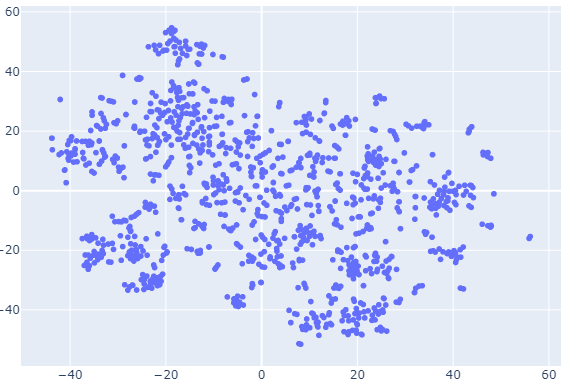}
        \caption{DeepWalk + t-SNE}
        \label{fig:bb}
    \end{subfigure}
    \caption{Graph layouts of \textsc{EmailUniv} dataset using SPLEE combined with t-SNE, compared with the baseline combination of DeepWalk and t-SNE.}
    \label{fig:email}
\end{figure}

\paragraph{Clustering accuracy.}
The NMI scores are used to test clustering accuracy of graph visualization.
The combination of SPLEE with t-SNE shows the best quality performance in all three datasets.
The baseline combination of DeepWalk and t-SNE also gives good results, but not as good as the previous combination.
The ascendance of SPLEE combined with t-SNE is most clearly shown in the \textsc{EmailUniv} dataset, as can be seen in Figure~\ref{fig:email}.
$GE$ methods combined with t-SGNE fall below the baseline, but they are almost as good as the baseline combination, which is still pleasing with regard to quality.
It is worth noticing that when the cluster structures in the original graph are clear (e.g. \textsc{EmailUniv} and \textsc{Lastfm}), different $GE$ methods give completely the same layout combined with t-SGNE.
This may be because t-SGNE takes both the high dimensional embedding and the original graph as inputs, which reduces the impact of the embedding to the layout.
Another remarkable observation is that t-SGNE largely improves the NMI performance of ShortestPath.
Combined with t-SNE, ShortestPath falls far behind the other two $GE$ methods, but combined with t-SGNE, its performance is almost as good as the other two.

\begin{figure}[htb]
    \centering
    \begin{subfigure}{0.3\textwidth}
        \includegraphics[width=\linewidth]{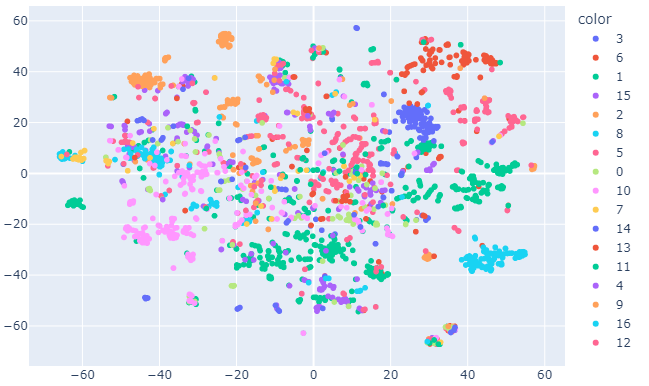}
        \caption{DeepWalk + t-SNE}
        \label{fig:a}
    \end{subfigure}\hfil
    \begin{subfigure}{0.3\textwidth}
        \includegraphics[width=\linewidth]{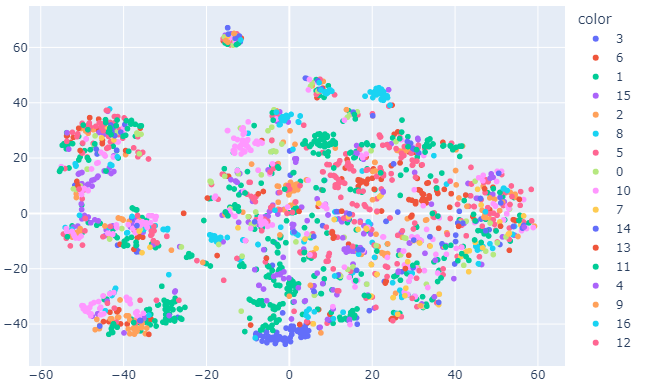}
        \caption{ShortestPath + t-SNE}
        \label{fig:b}
    \end{subfigure}\hfil
    \begin{subfigure}{0.3\textwidth}
        \includegraphics[width=\linewidth]{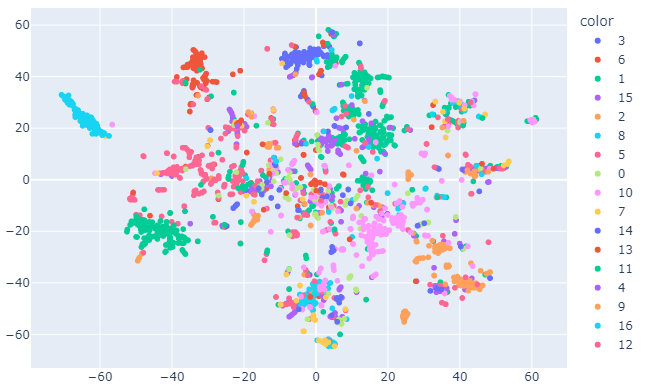}
        \caption{SPLEE + t-SNE}
        \label{fig:c}
    \end{subfigure}
    \medskip
    \begin{subfigure}{0.3\textwidth}
        \includegraphics[width=\linewidth]{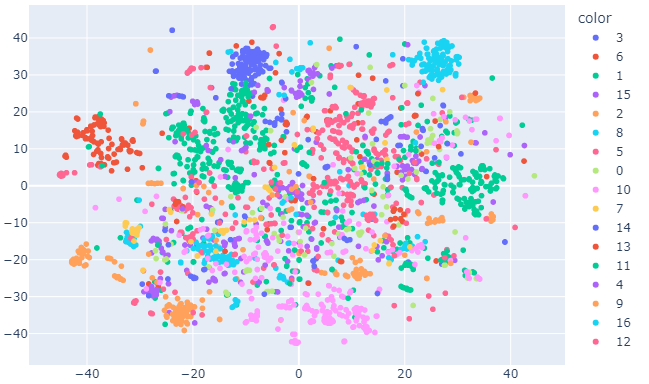}
        \caption{DeepWalk + t-SGNE}
        \label{fig:d}
    \end{subfigure}\hfil
    \begin{subfigure}{0.3\textwidth}
        \includegraphics[width=\linewidth]{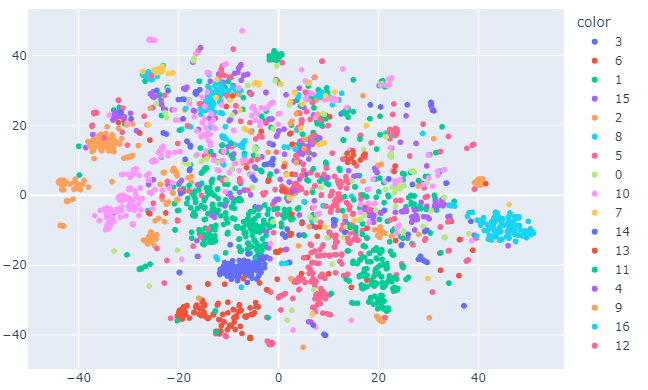}
        \caption{ShortestPath + t-SGNE}
        \label{fig:e}
    \end{subfigure}\hfil
    \begin{subfigure}{0.3\textwidth}
        \includegraphics[width=\linewidth]{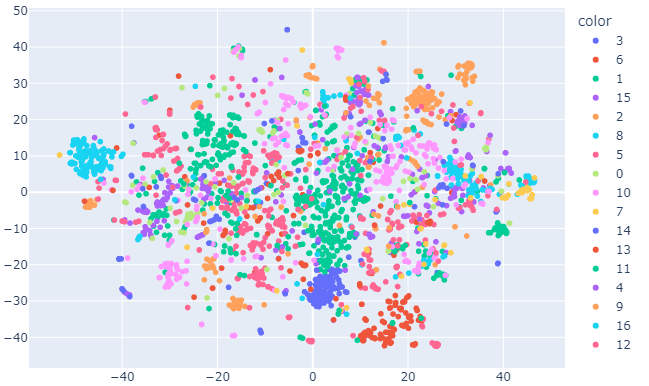}
        \caption{SPLEE + t-SGNE}
        \label{fig:f}
    \end{subfigure}
    \caption{Graph layouts of \textsc{Wiki} dataset using different combinations of $GE$ and $DR$ methods.}
    \label{fig:wiki}
\end{figure}

\paragraph{Aesthetics quality.}
From the results, we can see that the AQ scores almost coincide with NMI scores, with the combination of SPLEE and t-SNE still taking the lead.
Different $GE$ methods combined with t-SGNE perform as well as this combination in some datasets (e.g. \textsc{EmailUniv}) and less satisfying in others, but overall their performances are pleasing for application.
We also directly compare the layouts of these different combinations visually.
Here we pick the \textsc{Wiki} dataset as is shown in Figure~\ref{fig:wiki}, since its cluster structures are ambiguous, so even slight differences in aesthetic performances can be clearly observed.
SPLEE combined with t-SNE (Figure~\ref{fig:c}) gives the best layout of \textsc{Wiki}, separating the cluster structures the most clearly among all combinations.
As a baseline, DeepWalk combined with t-SNE (Figure~\ref{fig:a}) separates some of the clusters on the periphery, but the borders are not as clear.
When using t-SGNE as the $DR$ method (Figure~\ref{fig:d}, \ref{fig:e}, \ref{fig:f}), clusters are separated, but the clusters seem bloated and mixed together with no clear borders.
Though below baseline, the results of t-SGNE are still acceptable since cluster structures can indeed be distinguished visually, and the such a problem will be covered up in graphs with clearer cluster structures (e.g. \textsc{EmailUniv}), as can be seen in Table~\ref{exp_01} from the AQ scores.

Hence, we can conclude that SPLEE combined with t-SNE produces layouts of the best quality.
However, we also notice that SPLEE cannot deal with large datasets decently due to quadratic running time.
Considering larger datasets, ShortestPath combined with t-SGNE is the best alternative choice.
In the next experiment on large datasets, ShortestPath and t-SGNE will be the selected pair of $GE$ and $DR$ methods.

%% file: Sections/Experiments/02.tex
In this experiment, we compare the runtime of t-SNE and t-SGNE.
Based on the results of Experiment~\ref{experiment1}, we fix our $GE$ method to be ShortestPath and apply t-SNE and t-SGNE respetively on synthetic datasets of increasing size to compare their runtime.
The datasets are generated using Lancichinetti–Fortunato–Radicchi (LFR) benchmark, which is an algorithm for generating synthetic networks with adjustable cluster structure~\cite{lancichinetti_benchmarks_2009}.
LFR benchmark is typically used to compare between differnet community detection algorithm.
It can generate datasets of arbitrary size with \textit{a priori} known clusters.
The data generation involves three parameters~\cite{lancichinetti_benchmarks_2009}:
\begin{itemize}
    \item $\tau_1$: power law exponent of the degree distribution of the graph;
    \item $\tau_2$: power law exponent of the community size distribution of the graph;
    \item $\mu\in[0,1]$: fraction of inter-community edges incident to each node (inter-community degree is $\mu\cdot\mathrm{deg}(u)$).
    Smaller $\mu$ values typically correspond to graphs with more separated clusters.
\end{itemize}
In this experiment, we use the implementation of LFR benchmark in the \texttt{scikit-learn} library of Python.
We set $\tau_1 = 4$, $\tau_2 = 4$ and $\mu = 0.18$.
The parameters are tuned so that the generated graphs will have clearly distinguishable cluster structures.
Table~\ref{exp02_cmp} shows the results of this experiment\footnote{During this experiment, there were other processes running on the same machine. To avoid their influences on the results, we perform each experiment three times and take the average.}.

\begin{table}[htb]
    \centering
    \begin{tabular}{|c|c|c|c|}
        \hline
        \textbf{Dataset} & \textbf{Size (Nodes / Edges)} & \textbf{ShortestPath / t-SNE / Total} & \textbf{ShortestPath / t-SGNE / Total} \\
        \hline\
        \textsc{Lfr\_30k\_0.18} & 30,000 / 75,643 &  0:00:04 / 0:01:33 / 0:01:37  & 0:00:04 / \textbf{0:00:23} / \textbf{0:00:27} \\
        \textsc{Lfr\_100k\_0.18} & 100,000 / 253,036 & 0:00:17 / 0:08:12 / 0:08:29 & 0:00:17 / \textbf{0:01:21} / \textbf{0:01:39}  \\
        \textsc{Lfr\_300k\_0.18} & 300,000 / 756,664 & 0:00:57 / 0:32:43 / 0:33:40 & 0:01:03 / \textbf{0:03:54} / \textbf{0:04:57} \\
        \hline
    \end{tabular} \\ \vskip .3cm
    \caption{Runtime comparison of t-SNE and t-SGNE on LFR-generated large datasets in the format of \texttt{h:mm:ss}. The machine we use for this experiment has a 32-core Intel Xeon Gold 6338 Processor CPU and 16GB of memory .}
    \label{exp02_cmp}
\end{table}

The results of this experiment are in accordance with our analysis in Section~\ref{method}.
The linear-time construction of the neighborhood graph in t-SGNE significantly reduces the running time as the size of the graph increases, compared with t-SNE in which the construction runs quadratic time.
Hence, t-SGNE is a more suitable algorithm for visualizing large graph datasets.

%% file: Sections/Experiments/03.tex
In this experiment, we further push the limit of the combination of ShortestPath and t-SGNE to larger graph datasets.
Based on the result of Experiment~\ref{experiment1} and Experiment~\ref{experiment2}, this combination of $GE$ and $DR$ methods has the best runtime performance and a satisfying performance in the quality of the layout.
We apply this combination to graph datasets involving both generated and real-world data, with sizes ranging from 10K to 4M as follows:

\begin{itemize}
    \item \textsc{Lfr\_30k\_0.18}, \textsc{Lfr\_300k\_0.18}, and \textsc{Lfr\_3m\_0.18} are datasets generated by the LFR benchmark as in Experiment~\ref{experiment2}, with 30K, 300K, and 3M nodes respectively.
    
    \item \textsc{TwitchGamers}~\cite{rozemberczki_twitch_2021} is a network of Twitch users, where the nodes represent the users and the edges represent their following relationships. It has 168,114 nodes and 6,797,557 edges.
    
    \item \textsc{Dblp}~\cite{yang_defining_2012} is a co-authorship network of researchers publishing papers of Computer Science. Nodes represent the authors and a link exists if two authors have at least one publication in common. The network has 317,080 nodes and 1,049,866 edges.
    
    \item \textsc{YoutubeComm}~\cite{yang_defining_2012} is a network of YouTube users, where the nodes represent the users and the edges represent the online friendship. It consists of 1,134,890 nodes and 2,987,624 edges, with 8385 user-defined communities.
    
    \item \textsc{LiveJournal}~\cite{leskovec_community_2008} is a network of LiveJournal users. Nodes represent the users and a link exists if a user declare the other member as a friend. The network has 3,997,962 nodes and 34,681,189 edges.
\end{itemize}

\begin{table}[htb]
    \centering
    \begin{tabular}{|c|c|c|c|c|}
        \hline
        \textbf{Type} & \textbf{Dataset} & \textbf{Size (Nodes / Edges)} & \textbf{ShortestPath} & \textbf{t-SGNE} \\
        \hline
        \multirow{3}{*}{Generated} & \textsc{Lfr\_30k\_0.18} & 30,000 / 75,643  & 0:00:02 & 0:00:13 \\
        & \textsc{Lfr\_300k\_0.18} & 300,000 / 756,664  & 0:00:28 & 0:02:31 \\
        & \textsc{Lfr\_3m\_0.18} & 3,000,000 / 4,937,941 & 0:05:26 & 0:49:02 \\
        \hline
        \multirow{4}{*}{Real-world} & \textsc{TwitchGamers} & 168,114 / 6,797,557 & 0:01:18 & 0:24:34 \\
        & \textsc{Dblp} & 317,080 / 1,049,866 & 0:00:27 & 0:12:33 \\
        & \textsc{YoutubeComm} & 1,134,890 / 2,987,624 & 0:02:54 & 3:09:23 \\
        & \textsc{LiveJournal} & 3,997,962 / 34,681,189 & 0:45:11 & 2:05:51 \\
        \hline
    \end{tabular} \\ \vskip .3cm
    \caption{Runtime of ShortestPath combined with t-SGNE on large datasets in the format of \texttt{h:mm:ss}. The machine we use for the experiment has an 8-core Apple M2 CPU and 16 GB of memory.}
    \label{exp03_cmp}
\end{table}

As can be seen from the results in Table~\ref{exp03_cmp}, the runtime of ShortestPath combined with t-SGNE is overall satisfying on large graph datasets, finishing in less than a few hours as the size of the graph scales up to 4M nodes.
No memory leak occurred at the million scale either.
We do notice that the runtime of t-SGNE involves randomness in certain real-world datasets (e.g. \textsc{YoutubeComm}).
This may be due to randomness added to the stochastic machine learning algorithms, which guarantees different models for each training.
However, such an abnormal increase in the runtime will not exceed an acceptable scale, since the maximum number of iterations is limited to a constant during the learning process.

%% file: Sections/4_Discussion.tex
In this section, we briefly discuss some problems and deficiencies of our currently proposed methods and the potential future improvements to them.

\paragraph{General types of graphs.}
Although we restrict our discussion to undirected, unweighted graphs with no node attributes, t-SGNE can be applied in a more general setting.
For weighted graphs, one possible modification is to construct a weighted neighborhood graph $G_{knn,G}$ from $G$, where the weight  of the edge from $x_i$ to $x_j$ is the average of product of weights on the path of random walks from $x_i$ to $x_j$.
One can also come up with a more sophisticated function of the weights on the path to avoid the potential quick decay of $w_{ij}$ as $k$ grows.

\paragraph{Other $GE$ methods.}
The reason why t-SGNE is not suitable for $GE$ methods like SDNE is that $G_{knn,X}$ cannot approximate the distance structures in $X$.
This raises a natural question: can we perform graph exploring algorithms more sophisticated than BFS, or more sophisticated analyses on the result of random walks, so that we can capture the more complex distance structure?
For example, maybe a comparison of the degrees can capture the structural similarity of nodes.

\paragraph{Other combinations of $GE$ and $DR$ methods.}
ShortestPath combined with t-SGNE is currently the selected combination for visualizing large graphs due to its pleasing visualization quality and reasonable runtime.
However, SPLEE combined with t-SNE is actually the best combination considering visualization quality.
Is it possible to improve the quality of the combination of ShortestPath and t-SGNE, or reduce the runtime of the combination of SPLEE and t-SNE, in order to obtain a better layout in reasonable time?